\documentclass[times,mathptm,sigconf,screen,authorversion]{acmart}
\usepackage{times}
\usepackage{latexsym}
\usepackage{graphicx}
\usepackage{paralist} % inparaenum
\usepackage{color}

\usepackage{amsmath,bm}

\usepackage{cleveref}
\usepackage{amssymb}
\usepackage{multirow}
\usepackage{caption}
\usepackage{booktabs}
\usepackage{tikz}
\usepackage{pbox}
\usetikzlibrary{patterns}
\usepackage{pgf,pgfplots}
\newcommand{\rpm}{\raisebox{.2ex}{$\scriptstyle\pm$}}

\AtBeginDocument{%
  \providecommand\BibTeX{{%
    \normalfont B\kern-0.5em{\scshape i\kern-0.25em b}\kern-0.8em\TeX}}}

\copyrightyear{2021}
\acmYear{2021}
\setcopyright{acmlicensed}\acmConference[ICAIL'21]{Eighteenth International Conference for Artificial Intelligence and Law}{June 21--25, 2021}{S\={a}o Paulo, Brazil}
\acmBooktitle{Eighteenth International Conference for Artificial Intelligence and Law (ICAIL'21), June 21--25, 2021, S\={a}o Paulo, Brazil}
\acmPrice{15.00}
\acmDOI{10.1145/3462757.3466085}
\acmISBN{978-1-4503-8526-8/21/06}

\begin{document}

\title{ Structural Text Segmentation of Legal Documents }

\author{Dennis Aumiller}
\authornote{Both authors contributed equally to this research.}
\affiliation{%
  \institution{Institute of Computer Science, Heidelberg University}
  \city{Heidelberg}
  \country{Germany}}
\email{aumiller@informatik.uni-heidelberg.de}

\author{Satya Almasian}
\authornotemark[1]
\affiliation{%
  \institution{Institute of Computer Science, Heidelberg University}
  \city{Heidelberg}
  \country{Germany}}
\email{almasian@informatik.uni-heidelberg.de}

\author{Sebastian Lackner}
\affiliation{%
  \institution{Institute of Computer Science, Heidelberg University}
  \city{Heidelberg}
  \country{Germany}}
\email{lackner@informatik.uni-heidelberg.de}

\author{ Michael Gertz}
\affiliation{%
  \institution{Institute of Computer Science, Heidelberg University}
  \city{Heidelberg}
  \country{Germany}
}
\email{gertz@informatik.uni-heidelberg.de}

\renewcommand{\shortauthors}{Aumiller and Almasian, et al.}

%!TeX root=main_paper.tex

\begin{abstract}

The growing complexity of legal cases has lead to an increasing interest in legal information retrieval systems that can effectively satisfy user-specific information needs.
However, such downstream systems typically require documents to be properly formatted and segmented, which is often done with relatively simple pre-processing steps, disregarding topical coherence of segments.
Systems generally rely on representations of individual sentences or paragraphs, which may lack crucial context, or document-level representations, which are too long for meaningful search results.
To address this issue, we propose a segmentation system that can predict topical coherence of sequential text segments spanning several paragraphs, effectively segmenting a document and providing a more balanced representation for downstream applications.
We build our model on top of popular transformer networks and formulate structural text segmentation as topical change detection, by performing a series of independent classifications that allow for efficient fine-tuning on task-specific data.
We crawl a novel dataset consisting of roughly $74{,}000$ online Terms-of-Service documents, including hierarchical topic annotations, which we use for training. Results show that our proposed system significantly outperforms baselines, and adapts well to structural peculiarities of legal documents. We release both data and trained models to the research community for future work.\footnote{\url{https://github.com/dennlinger/TopicalChange}} 

%Mention how legal documents have domain-specific vocabulary and can be complex in structure and length.
%This leads to the natural question how such documents can be best represented for downstream applications, such as Information Retrieval (IR) or similarity search.
%Problem is that there are currently two extremes: Sentence-level approaches, which lack context, and document-level inputs, which cannot capture finer semantic changes.
%Therefore, it makes sense to have topically coherent segments instead, which are loosely based on the notion of ``paragraphs''
%Many texts already indicate topical coherence in the form of visually coherent segments, but this information is only infrequently used (maybe cite the Mecia paper?) in actual preprocessing.
%We investigate text segmentation on the paragraph level, specifically for legal documents.
%Our system shows that Transformer models can be adopted to recognize topical changes in documents, and experiment on a corpus of Terms-of-Service texts from Websites.

\end{abstract}

\begin{CCSXML}
<ccs2012>
<concept>
<concept_id>10010405.10010455.10010458</concept_id>
<concept_desc>Applied computing~Law</concept_desc>
<concept_significance>300</concept_significance>
</concept>
<concept>
<concept_id>10010405.10010497.10010510.10010513</concept_id>
<concept_desc>Applied computing~Annotation</concept_desc>
<concept_significance>500</concept_significance>
</concept>
<concept>
<concept_id>10002951.10003317.10003318.10003319</concept_id>
<concept_desc>Information systems~Document structure</concept_desc>
<concept_significance>500</concept_significance>
</concept>
</ccs2012>
\end{CCSXML}

\ccsdesc[300]{Applied computing~Law}
\ccsdesc[500]{Applied computing~Annotation}
\ccsdesc[500]{Information systems~Document structure}

\keywords{Document Understanding, Outline Generation, Text Segmentation}

\maketitle

\section{Introduction}
\label{sec:intro}
%!TeX root=main_paper.tex

% Introduction 
Written texts are often a sequence of semantically coherent segments, designed to create a smooth transition between various subtopics discussed in a single document. Usually, the information needs of a user are satisfied by retrieving only the relevant subtopic, and retrieving the whole document is unwieldy and may result in information overload~\cite{DBLP:journals/ipm/MisraYCJ11,DBLP:conf/sigir/Wilkinson94}. However, the context of a single subtopic frequently spans multiple sentences and contains localized context, which is crucial for proper understanding.
Despite the clear relevance of segmentation to downstream performance, many (legal) retrieval systems choose structurally rigid representations of only a single text element (generally either the full document \cite{DBLP:conf/icail/ConradAZK05, DBLP:conf/jurix/Mimouni13}, or a single paragraph/sentence \cite{DBLP:conf/icail/PoudyalGQ19, DBLP:conf/jurix/WestermannSWAB20}), disregarding the semantic coherence.
Especially in legal documents, which can be extremely lengthy and contain domain-specific complexities in their topics, it is important to suitably represent entire topics in a single cohesive unit.
Furthermore, aside from semi-structured legal texts, such as laws, other documents do not necessarily contain uniform and easily separable segments, to begin with. Especially input formats such as PDF or scans frequently lack any sort of meta descriptors for hierarchical information about the document contents, which makes this a challenging task.
To find a fitting representation that captures the precise topical context in the text, a robust and flexible framework to obtain such a structural segmentation is required. 
We therefore propose a new approach for the estimation of topic boundaries to generate more suitable document representations for the mentioned downstream applications, by considering the topical coherence of paragraphs. 
We define a coherent section in a document as a unit consisting of potentially one or multiple paragraphs, which together share a common topic. Section boundaries often coincide with a change in topic and can thus be assumed to generate candidates for the later segmentation.

\begin{figure}
\centering 
\resizebox{0.5\textwidth}{0.15\textwidth}{      
\input{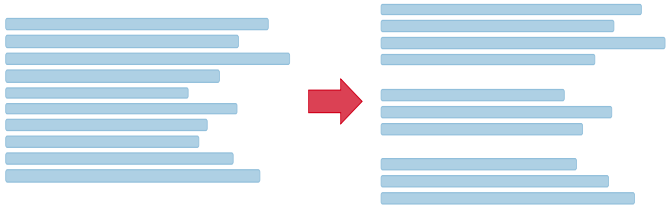}
}
\caption{Visual cues such as paragraphs often give away a notion of semantic coherence, which is disregarded in sentence-level models.}
\label{fig:paragraph}
\end{figure}

Despite their importance, many previous works for structural text segmentation ignore the notion of paragraphs and focus only on the granularity of sentences. This is contrary to the nature of written text, where paragraphs represent a semantically cohesive unit, which is already available and represents a coarser and more meaningful structure than sentences. 
In this work, we assume that topic boundaries generally do not appear in the middle of a paragraph, and, consequently, operating on paragraph level can reduce the risk of false-positive segmentations and lower the computation cost of per sentence prediction. Figure~\ref{fig:paragraph} shows how paragraphs group sentences and divide a text into coherent parts and how by overlooking this valuable information the structure in the text is lost to the model.

\noindent Focusing on the field of text segmentation in the context of the Natural Language Processing (NLP), we find an already large body of existing work.
Because no large labeled dataset existed, early approaches to text segmentation were mainly unsupervised, using heuristics to identify whether two sentences belong to the same topic or not. Such approaches either exploit the fact that topically related words tend to appear in semantically coherent segments~\cite{10.5555/974305.974309,10.5555/972684.972687,DBLP:conf/acl/MalioutovB06,DBLP:conf/acl/Kozima93,DBLP:conf/acl/UtiyamaI01}, or focus on the representation of text in terms of latent-topic vectors using topic modeling methods, such as Latent Dirichlet Allocation~\cite{DBLP:journals/jmlr/BleiNJ03,DBLP:journals/ipm/MisraYCJ11,DBLP:conf/cikm/MisraYJC09,DBLP:conf/naacl/RiedlB12}. Recently, with the availability of annotated data, text segmentation has also been formulated as a supervised learning problem. Most methods utilize a hierarchical neural model, where the lower-level network creates sentence representations, and a secondary network models the dependencies between the embedded sentences~\cite{DBLP:journals/corr/abs-2001-00891,DBLP:conf/naacl/KoshorekCMRB18}. These models use sentence dependencies to predict the potential segment boundaries. One drawback of these approaches is their sentence-level granularity, which disregards the paragraph coherence previously mentioned. This problem is partially solved by hierarchical neural models, where the dependency between sentences is modeled in a hierarchical structure, to combine sentence into bigger units. However, training such models are, due to the different document lengths, computationally expensive. Moreover, these models fail to take advantage of larger pre-trained language representations, such as BERT~\cite{DBLP:conf/naacl/DevlinCLT19} and RoBERTa~\cite{DBLP:journals/corr/abs-1907-11692}, which have recently proven to be valuable feature generators with a low cost of fine-tuning, advancing the state-of-the-art in several disciplines.

\noindent In this paper, we aim to tackle the task of structural text segmentation using transformer-based models, and introduce a novel dataset of Terms-of-Service documents, containing annotated paragraphs belonging to the same topic. We focus on the text segmentation alone and leave the use of segmentation for retrieval enhancement to future work. We formulate topical coherence as a special case of binary classification of Same Topic Prediction (STP) and fine-tune our transformer-based models to detect paragraphs belonging to sections with similar topics. We consider sections as the top-level hierarchy in our dataset and do not consider subsections.
We assume topical independence between consecutive paragraphs and show that it does not lead to a deterioration of performance while avoiding the costly computation of hierarchical models. 
The hypothesis is that by fine-tuning the paragraph embedding for topic similarity we can generate segment features that detect coherent topical structures in a document. We evaluate our models against the traditional embedding baselines and compare them to supervised and unsupervised approaches for text segmentation and find significant improvements by our method.\\
\\
\noindent
\textbf{Contributions.}
The contributions of this paper are as follows:
\begin{inparaenum}[$(i)$]
\item We present the task of structural text segmentation on coarser cohesive text units (paragraphs/sections).
\item We investigate the performance of transformer-based models for topical change detection, and
\item frame the task as a collection of independent binary predictions, reducing overhead for hierarchical training and simplified training sample generation.
\item We present a new dataset consisting of online Terms-of-Service documents partitioned into hierarchical sections, and make the data available for future research.
\item We evaluate our model against classical baselines for text segmentation, and
\item show the effectiveness of our generated embeddings for structural segmentations to obtain superior performance to other text segmentation techniques.
\end{inparaenum}

\section{Related Work}
\label{sec:related}
%!TeX root=main_paper.tex

Our work is closely tied to the broader fields of legal document understanding, topic analysis, text segmentation, and transformer language models, and we briefly review related work in each of these areas in this section. 

\subsection{Legal Document Understanding}
The area of legal document understanding and legal information retrieval has a long-standing history. A great overview is presented by Moens~\cite{DBLP:journals/ail/Moens01}, who details more of the applications that were mentioned in the introduction. While there is existing work on the topic of legal document segmentation \cite{DBLP:conf/icail/Mencia09, DBLP:conf/cikm/LuCAK11, DBLP:conf/icail/LyteB19}, they are generally concerned with a specific information extraction interest. In the case of Menc\'ia, they are concerned with metadata of French law documents \cite{DBLP:conf/icail/Mencia09}. Specifically, they also make use of an existing HTML/XML structure in their input documents, but do not generalize to arbitrary text inputs without structural features.
Lu et al., on the other hand, are utilizing clustering techniques to identify sub-topics in legal documents. However, these topics are irrespective of actual section boundaries within the document~\cite{DBLP:conf/cikm/LuCAK11}. They further include additional metadata, such as citations, headnotes, and key numbers, in their task setup, which are suitable to their specific application in case law.
Similarly, Conrad et al.~\cite{DBLP:conf/icail/ConradAZK05} have previously attempted to employ clustering for heterogeneous document collections, again focusing on hierarchical representations in the form of short topic descriptors and not focusing on the actual textual content of the documents themselves.
Lyte and Branting~\cite{DBLP:conf/icail/LyteB19} classify metadata labels based on a CRF (Conditional Random Field) model, building on prior work by Branting~\cite{DBLP:conf/icail/Branting17} in the same direction, but focus mostly on element classification.
Slightly longer segments in the form of entire sentences are both used by Poudyal et al.~\cite{DBLP:conf/icail/PoudyalGQ19}, who mine arguments from European case-law decision, and Westermann et al.~\cite{DBLP:conf/jurix/WestermannSWAB20}, where a system for efficient similarity search based on sentence embeddings is presented.

\subsection{Topic Analysis}
Detection and analysis of topical change are grounded in topic modeling approaches. Earlier work such as LDA~\cite{DBLP:journals/jmlr/BleiNJ03} treat documents as bag-of-words, where each document is assigned to a topic distribution, and each topic is a distribution over all words. More recent work has adopted a more sophisticated representation than bag-of-words and generally models Markovian topic or state transitions to capture dependencies between words in a document~\cite{DBLP:conf/nips/GriffithsSBT04,DBLP:conf/icml/Wallach06}. With the rise of distributed word representation, the focus has shifted to the combination of LDA and word embeddings~\cite{DBLP:journals/corr/abs-1907-04907,DBLP:journals/corr/Moody16}. Since we are interested in a primary segmentation without necessarily predicting topics, we put a stronger focus on the related work of segmentation methods, as discussed in the following section.

\subsection{Text Segmentation}
Text Segmentation is the task of dividing a document into a multi-paragraph discourse unit that is topically coherent, with the cut-off point usually indicating a change in topic~\cite{DBLP:conf/acl/Hearst94,DBLP:conf/acl/UtiyamaI01}. Although the task itself dates back to 1994~\cite{DBLP:conf/acl/Hearst94}, most existing text segmentation datasets are small and limit their scope to sentences (predicting whether two sentences discuss the same topic or not). The most common dataset is by Choi~\cite{DBLP:conf/anlp/Choi00}, containing only $920$ synthesized passages from the Brown corpus. Choi's method (C99) is a probabilistic algorithm measuring similarity via term overlap. GraphSeg~\cite{DBLP:conf/starsem/GlavasNP16} is an unsupervised graph method that segments documents using a semantic relatedness graph of a document. 
GraphSeg is also evaluated on a small set of $5$ manually-segmented political manifestos from the Manifesto project\footnote{\url{https://manifestoproject.wzb.eu}}. 
Another class of methods are topic-based document segmentations, which are statistical models that find latent topic assignments reflecting the underlying structure of the document~\cite{DBLP:journals/ml/BeefermanBL99,DBLP:conf/cikm/BrantsCT02,DBLP:conf/naacl/ChenBBK09,DBLP:conf/interspeech/DharanipragadaFMRW99,DBLP:journals/ipm/MisraYCJ11,riedl12_acl}.
TopicTiling~\cite{riedl12_acl} performs best among this family of methods and uses LDA to detect topic shifts, with computing similarities between adjacent blocks based on their term frequency vectors.
Brants et al.~\cite{DBLP:conf/cikm/BrantsCT02} follow a similar approach but employ PLSA~\cite{DBLP:journals/sigir/Hofmann17} to compute the estimated word distributions. Another noteworthy approach based on Bayesian topic models is by Chen et al.~\cite{DBLP:conf/naacl/ChenBBK09}, where they constrain latent topic assignments to reflect the underlying organization of document topics. They also publish a test dataset with $218$ Wikipedia articles about cities and chemical elements.

\noindent All mentioned methods are unsupervised learning approaches, and small annotated datasets are only used for evaluation and, hence, are not directly comparable to our approach. Instead, we focus on supervised learning of topics and introduce a new dataset with $43{,}056$ automatically labeled documents.

\noindent The only two comparable supervised approaches are from Koshorek et al.~\cite{DBLP:conf/naacl/KoshorekCMRB18} and Glavas et al.~\cite{DBLP:journals/corr/abs-2001-00891}. Koshorek et al.~\cite{DBLP:conf/naacl/KoshorekCMRB18} propose a hierarchical LSTM architecture for learning sentence representation and their dependencies. They train their hierarchical model on a dataset of cleaned Wikipedia articles, called Wiki-$727$k. Glavas et al.~\cite{DBLP:journals/corr/abs-2001-00891} introduce Coherence-Aware Text Segmentation, which encodes a sentence sequence using two hierarchically connected transformer networks. The two latter models are closest to our work in terms of data size and problem formulation. However, they rely solely on per sentence predictions, which is incomparable to our paragraph-based method. The model by Glavas et al.~is similar to our approach in that it is based on a transformer architecture, yet, they do not take advantage of transfer learning from pre-trained language models and learn all the features from scratch. Finally, Zhang et al.~\cite{DBLP:conf/sigir/ZhangGFLC19} extend text segmentation by outline generation and trained an end-to-end LSTM-model for identifying sections and generating corresponding headings for Wikipedia documents. 

\subsection{Transformer Language Models}
\sloppy
The transformer architecture, much like recurrent neural networks, aims to solve sequence-to-sequence tasks, relying entirely on self-attention to compute representations of its input and output~\cite{DBLP:conf/nips/VaswaniSPUJGKP17}.
Transformers have made a significant step in bringing transfer learning to the NLP community, which allows the easy adaptation of a generically pre-trained model for specific tasks. Pre-trained models such as BERT, GPT-2, and RoBERTa~\cite{DBLP:conf/naacl/DevlinCLT19,DBLP:journals/corr/abs-1907-11692,RadfordWCLAS} use language modeling for pre-training on large corpora.
These models are powerful feature generators, which with minimal task-specific fine-tuning achieve state-of-the-art performance on a wide variety of tasks. Although at the core of all these models lies the idea of transformers and attention mechanisms, many have been modified and optimized to fit various downstream applications. One variation based on BERT is Sentence-BERT~\cite{DBLP:conf/emnlp/ReimersG19}, which combines two BERT-based models in Siamese fashion to derive semantically meaningful sentence embeddings. 
By its design, Sentence-BERT also allows for longer input sequences for pairwise training tasks and outperforms BERT on semantic textual similarity tasks, making it a suitable choice for embedding paragraphs.
Another notable variant of BERT is RoBERTa, a retraining of BERT with improved training methodology and more training data, it achieves slightly better results than BERT on some natural understanding tasks. Due to the advantages of  RoBERTa, we chose RoBERTa and Sentence-RoBERTa from the Sentence-BERT variant for the setup in our approach.

\section{Same Topic Prediction}
\label{sec:model}
%!TeX root=main_paper.tex

We formulate structural text segmentation as a supervised learning task of the same topic prediction. Our model consists of two steps: \begin{inparaenum}[$(i)$]
\item Independent and Identically Distributed Same Topic Prediction (IID STP) and
\item Sequential inference over a full document. 
\end{inparaenum}
As mentioned previously, sections are the considered level of hierarchy in our model and the structure of sub-sections is ignored in this study. However, the model is easily adaptable to any granularity, and our dataset contains information for all the levels.
In the first step, we fine-tune transformer-based models to detect topical change for both paragraphs and entire sections. Given two paragraphs or sections, the classifier should correctly identify if they discuss the same subject or not. We assume that the topic of each paragraph or section is independent of the text before and after, meaning that the topic of one paragraph does not affect the likelihood of the next paragraph belonging to the same topic. We later prove that this assumption yields good performance without a costly training of hierarchical models. In the second step, we use the fine-tuned transformer-based classifiers for sequential inference on entire documents, where the segment boundaries are defined by topical change. In the following, we discuss these steps in more detail. 

\subsection{IID Same Topic Prediction (STP)}\label{sec:STP}
A document $d\in D$ is represented as a sequence of $N$ sections $S_{d}=(s_{1},..., s_{N})$, where each section is assigned one of $M$ topics $T=(t_{1},...,t_{M})$, and 
each section contains up to $K$ paragraphs $P_{n}=(p_{1},....p_{K})$. We assume topical consistency within a paragraph and argue that the results for classification do not change based on the position of the paragraph in the document, since the most relevant part for our inference is the intra-section information. Therefore, all paragraphs in a section belong to the same topic. If the topic assignment is defined by the function $Topic$, we have: 
\begin{align}
&s_{n} =( p_{1},.., p_{k}) \land Topic (s_{n})=t_{1} \implies \nonumber\\  &Topic (p_{1})=t_{1} = ...= Topic (p_{k})=t_{1}
\end{align}
 If we define $C$ as a chunk of text corresponding to either a section or paragraph, the topic prediction task is defined for section and paragraph granularity as follows: Given two chunks of text of the same type (both paragraphs or both sections) $(c_{1}, c_{2})$ and labels $ y \in \{0,1\}$, indicating whether the two chunks belong to the same topic, topical change detection can be formulated as a binary classification problem. The positive class indicates that both chunks have the same topic, whereas the negative class indicates a change in topic and potentially the beginning of a new segment in text.
Note that we only consider chunks of the same type, namely, either \textit{only} sections or \textit{only} paragraphs, in each model. By formulating the problem as a binary classification, detecting the topic consistency between two chunks of text can now be solved with any type of classifier. In this work, we train two types of transformer-based classifiers for this task, one from the pre-trained language models~\cite{DBLP:journals/corr/abs-1907-11692} and another Siamese network~\cite{DBLP:conf/emnlp/ReimersG19} variation, which is more suitable for encoding pairwise similarity. Subsequently, the two variations are discussed.

\subsubsection{RoBERTa\nopunct} is a replication study of BERT pre-training with optimized hyper-parameters that applies minor adjustments to the BERT language model to achieve better performance~\cite{DBLP:journals/corr/abs-1907-11692}. BERT and RoBERTa both belong to the family of pre-trained transformer-based language models. The transformer is an architecture for shaping one sequence into another one with the help of the self-attention mechanism, which helps the model to extract features from each word relative to all the other words in the sequence. The encoder stacks in BERT and RoBERTa consist of one or multiple self-attention blocks followed by a feed-forward network. During pre-training, two sentences are taken as input, and models are trained on two tasks of language modeling, by predicting masked words in the input and next sentence prediction, and by classifying whether the two sentences are sequential. By these means, the models learn task-independent features from a vast amount of unlabelled text that can then be used in a fine-tuning stage for various natural language understanding tasks. 
Since the performance difference between most transformer-based language models is negligible, we choose RoBERTa as the representative of this family. 
In the fine-tuning process, the model receives two chunks as input and learns to predict whether they belong to the same topic or not. To distinguish between two chunks in training a \texttt{[CLS]} token is inserted at the beginning of the first chunk and a \texttt{[SEP]} token is inserted at the end of both the first and second chunk. The embedding of the \texttt{[CLS]} is what is used for pre-training the next sentence prediction task and contains RoBERTa's understanding at the sentence-level. This token is used by a simple classification layer, learned during fine-tuning, for the same topic prediction task.
 Since the input size for both chunks combined is limited to a maximum of $512$ tokens, shorter than many sections and paragraphs in our dataset, any longer chunk of text has to be truncated to fit. 
 
\subsubsection{Sentence-Transformers (SRoBERTa)\nopunct} aims to enhance the sentence embeddings by modification of RoBERTa using a Siamese architecture to derive semantically meaningful sentence embeddings~\cite{DBLP:conf/emnlp/ReimersG19}. Their method is available for several transformer models. We choose a RoBERTa-based variant to make the results comparable to the first approach. SRoBERTa enables RoBERTa to be used for certain new tasks, such as large-scale semantic similarity comparison. Their modifications result in faster inference and better representation for sentence-pair tasks. Moreover, because of the Siamese structure and coupling of two RoBERTa networks, the input size doubles, which allows for longer sequences and thus more context. In this setup, each sentence is passed through a separate RoBERTa network with an input limit of $512$ tokens. The sentence embeddings are derived from a pooling operation over the output of two models with tied weights. Sentence-Transformers introduce several learning objectives, out of which we use the classification objective function with binary cross-entropy loss to classify the chunks into the same topics. 

\subsection{Sequential Inference}
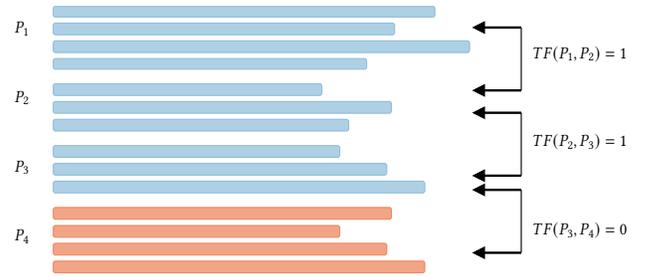
\begin{figure}
\centering 
\resizebox{0.47\textwidth}{0.2\textwidth}{      
\tikzset{every picture/.style={line width=0.75pt}} %set default line width to 0.75pt        

\begin{tikzpicture}[x=0.75pt,y=0.75pt,yscale=-1,xscale=1]
%uncomment if require: \path (0,323); %set diagram left start at 0, and has height of 323

%Rounded Rect [id:dp986548566125343] 
\draw  [color={rgb, 255:red, 145; green, 191; blue, 219 }  ,draw opacity=1 ][fill={rgb, 255:red, 145; green, 191; blue, 219 }  ,fill opacity=0.73 ] (47.8,18.8) .. controls (47.8,17.81) and (48.61,17) .. (49.6,17) -- (387,17) .. controls (387.99,17) and (388.8,17.81) .. (388.8,18.8) -- (388.8,24.2) .. controls (388.8,25.19) and (387.99,26) .. (387,26) -- (49.6,26) .. controls (48.61,26) and (47.8,25.19) .. (47.8,24.2) -- cycle ;
%Rounded Rect [id:dp5262265957786072] 
\draw  [color={rgb, 255:red, 145; green, 191; blue, 219 }  ,draw opacity=1 ][fill={rgb, 255:red, 145; green, 191; blue, 219 }  ,fill opacity=0.73 ] (47.8,33.22) .. controls (47.8,32.12) and (48.7,31.22) .. (49.8,31.22) -- (350.8,31.22) .. controls (351.9,31.22) and (352.8,32.12) .. (352.8,33.22) -- (352.8,39.22) .. controls (352.8,40.32) and (351.9,41.22) .. (350.8,41.22) -- (49.8,41.22) .. controls (48.7,41.22) and (47.8,40.32) .. (47.8,39.22) -- cycle ;
%Rounded Rect [id:dp019074975910590375] 
\draw  [color={rgb, 255:red, 145; green, 191; blue, 219 }  ,draw opacity=1 ][fill={rgb, 255:red, 145; green, 191; blue, 219 }  ,fill opacity=0.73 ] (47.8,48.44) .. controls (47.8,47.34) and (48.7,46.44) .. (49.8,46.44) -- (417.8,46.44) .. controls (418.9,46.44) and (419.8,47.34) .. (419.8,48.44) -- (419.8,54.44) .. controls (419.8,55.54) and (418.9,56.44) .. (417.8,56.44) -- (49.8,56.44) .. controls (48.7,56.44) and (47.8,55.54) .. (47.8,54.44) -- cycle ;
%Rounded Rect [id:dp35864160657274] 
\draw  [color={rgb, 255:red, 145; green, 191; blue, 219 }  ,draw opacity=1 ][fill={rgb, 255:red, 145; green, 191; blue, 219 }  ,fill opacity=0.73 ] (47.8,63.46) .. controls (47.8,62.47) and (48.61,61.66) .. (49.6,61.66) -- (326,61.66) .. controls (326.99,61.66) and (327.8,62.47) .. (327.8,63.46) -- (327.8,68.86) .. controls (327.8,69.85) and (326.99,70.66) .. (326,70.66) -- (49.6,70.66) .. controls (48.61,70.66) and (47.8,69.85) .. (47.8,68.86) -- cycle ;

%Rounded Rect [id:dp8458798119626911] 
\draw  [color={rgb, 255:red, 145; green, 191; blue, 219 }  ,draw opacity=1 ][fill={rgb, 255:red, 145; green, 191; blue, 219 }  ,fill opacity=0.73 ] (47.8,85.17) .. controls (47.8,84.07) and (48.7,83.17) .. (49.8,83.17) -- (285.8,83.17) .. controls (286.9,83.17) and (287.8,84.07) .. (287.8,85.17) -- (287.8,91.17) .. controls (287.8,92.27) and (286.9,93.17) .. (285.8,93.17) -- (49.8,93.17) .. controls (48.7,93.17) and (47.8,92.27) .. (47.8,91.17) -- cycle ;
%Rounded Rect [id:dp7285959214747992] 
\draw  [color={rgb, 255:red, 145; green, 191; blue, 219 }  ,draw opacity=1 ][fill={rgb, 255:red, 145; green, 191; blue, 219 }  ,fill opacity=0.73 ] (47.8,100.39) .. controls (47.8,99.29) and (48.7,98.39) .. (49.8,98.39) -- (348,98.39) .. controls (349.1,98.39) and (350,99.29) .. (350,100.39) -- (350,106.39) .. controls (350,107.49) and (349.1,108.39) .. (348,108.39) -- (49.8,108.39) .. controls (48.7,108.39) and (47.8,107.49) .. (47.8,106.39) -- cycle ;

%Rounded Rect [id:dp8601375949432721] 
\draw  [color={rgb, 255:red, 145; green, 191; blue, 219 }  ,draw opacity=1 ][fill={rgb, 255:red, 145; green, 191; blue, 219 }  ,fill opacity=0.73 ] (47.8,115.61) .. controls (47.8,114.51) and (48.7,113.61) .. (49.8,113.61) -- (309.8,113.61) .. controls (310.9,113.61) and (311.8,114.51) .. (311.8,115.61) -- (311.8,121.61) .. controls (311.8,122.71) and (310.9,123.61) .. (309.8,123.61) -- (49.8,123.61) .. controls (48.7,123.61) and (47.8,122.71) .. (47.8,121.61) -- cycle ;

%Rounded Rect [id:dp7094906867577653] 
\draw  [color={rgb, 255:red, 145; green, 191; blue, 219 }  ,draw opacity=1 ][fill={rgb, 255:red, 145; green, 191; blue, 219 }  ,fill opacity=0.73 ] (47.8,138.12) .. controls (47.8,137.02) and (48.7,136.12) .. (49.8,136.12) -- (301.8,136.12) .. controls (302.9,136.12) and (303.8,137.02) .. (303.8,138.12) -- (303.8,144.12) .. controls (303.8,145.22) and (302.9,146.12) .. (301.8,146.12) -- (49.8,146.12) .. controls (48.7,146.12) and (47.8,145.22) .. (47.8,144.12) -- cycle ;
%Rounded Rect [id:dp672529480057297] 
\draw  [color={rgb, 255:red, 145; green, 191; blue, 219 }  ,draw opacity=1 ][fill={rgb, 255:red, 145; green, 191; blue, 219 }  ,fill opacity=0.73 ] (47.8,153.34) .. controls (47.8,152.24) and (48.7,151.34) .. (49.8,151.34) -- (343.8,151.34) .. controls (344.9,151.34) and (345.8,152.24) .. (345.8,153.34) -- (345.8,159.34) .. controls (345.8,160.44) and (344.9,161.34) .. (343.8,161.34) -- (49.8,161.34) .. controls (48.7,161.34) and (47.8,160.44) .. (47.8,159.34) -- cycle ;
%Rounded Rect [id:dp864269109812964] 
\draw  [color={rgb, 255:red, 145; green, 191; blue, 219 }  ,draw opacity=1 ][fill={rgb, 255:red, 145; green, 191; blue, 219 }  ,fill opacity=0.73 ] (47.8,168.58) .. controls (47.8,167.48) and (48.7,166.58) .. (49.8,166.58) -- (377.8,166.58) .. controls (378.9,166.58) and (379.8,167.48) .. (379.8,168.58) -- (379.8,174.58) .. controls (379.8,175.68) and (378.9,176.58) .. (377.8,176.58) -- (49.8,176.58) .. controls (48.7,176.58) and (47.8,175.68) .. (47.8,174.58) -- cycle ;

%Rounded Rect [id:dp5028712516956915] 
\draw  [color={rgb, 255:red, 239; green, 138; blue, 98 }  ,draw opacity=1 ][fill={rgb, 255:red, 239; green, 138; blue, 98 }  ,fill opacity=0.77 ] (47.8,206.54) .. controls (47.8,205.44) and (48.7,204.54) .. (49.8,204.54) -- (301.8,204.54) .. controls (302.9,204.54) and (303.8,205.44) .. (303.8,206.54) -- (303.8,212.54) .. controls (303.8,213.64) and (302.9,214.54) .. (301.8,214.54) -- (49.8,214.54) .. controls (48.7,214.54) and (47.8,213.64) .. (47.8,212.54) -- cycle ;
%Rounded Rect [id:dp6880276121050182] 
\draw  [color={rgb, 255:red, 239; green, 138; blue, 98 }  ,draw opacity=1 ][fill={rgb, 255:red, 239; green, 138; blue, 98 }  ,fill opacity=0.77 ] (47.8,221.76) .. controls (47.8,220.66) and (48.7,219.76) .. (49.8,219.76) -- (343.8,219.76) .. controls (344.9,219.76) and (345.8,220.66) .. (345.8,221.76) -- (345.8,227.76) .. controls (345.8,228.86) and (344.9,229.76) .. (343.8,229.76) -- (49.8,229.76) .. controls (48.7,229.76) and (47.8,228.86) .. (47.8,227.76) -- cycle ;
%Rounded Rect [id:dp7322899754689955] 
\draw  [color={rgb, 255:red, 239; green, 138; blue, 98 }  ,draw opacity=1 ][fill={rgb, 255:red, 239; green, 138; blue, 98 }  ,fill opacity=0.77 ] (47.8,237) .. controls (47.8,235.9) and (48.7,235) .. (49.8,235) -- (377.8,235) .. controls (378.9,235) and (379.8,235.9) .. (379.8,237) -- (379.8,243) .. controls (379.8,244.1) and (378.9,245) .. (377.8,245) -- (49.8,245) .. controls (48.7,245) and (47.8,244.1) .. (47.8,243) -- cycle ;

%Rounded Rect [id:dp9753435406288808] 
\draw  [color={rgb, 255:red, 239; green, 138; blue, 98 }  ,draw opacity=1 ][fill={rgb, 255:red, 239; green, 138; blue, 98 }  ,fill opacity=0.77 ] (47.8,191.1) .. controls (47.8,190) and (48.7,189.1) .. (49.8,189.1) -- (348,189.1) .. controls (349.1,189.1) and (350,190) .. (350,191.1) -- (350,197.1) .. controls (350,198.2) and (349.1,199.1) .. (348,199.1) -- (49.8,199.1) .. controls (48.7,199.1) and (47.8,198.2) .. (47.8,197.1) -- cycle ;

%Straight Lines [id:da9460032637419911] 
\draw    (466,35) -- (466,88.86) ;
%Straight Lines [id:da20308673953735057] 
\draw [line width=1.5]    (466,35) -- (426,35) ;
\draw [shift={(422,35)}, rotate = 360] [fill={rgb, 255:red, 0; green, 0; blue, 0 }  ][line width=0.08]  [draw opacity=0] (11.61,-5.58) -- (0,0) -- (11.61,5.58) -- cycle    ;
%Straight Lines [id:da490750758183911] 
\draw [line width=1.5]    (466,88.86) -- (426,88.86) ;
\draw [shift={(422,88.86)}, rotate = 360] [fill={rgb, 255:red, 0; green, 0; blue, 0 }  ][line width=0.08]  [draw opacity=0] (11.61,-5.58) -- (0,0) -- (11.61,5.58) -- cycle    ;

%Straight Lines [id:da834339836692134] 
\draw    (466,174) -- (466,227.86) ;
%Straight Lines [id:da8838974009024314] 
\draw [line width=1.5]    (466,174) -- (426,174) ;
\draw [shift={(422,174)}, rotate = 360] [fill={rgb, 255:red, 0; green, 0; blue, 0 }  ][line width=0.08]  [draw opacity=0] (11.61,-5.58) -- (0,0) -- (11.61,5.58) -- cycle    ;
%Straight Lines [id:da06269576021553047] 
\draw [line width=1.5]    (466,227.86) -- (426,227.86) ;
\draw [shift={(422,227.86)}, rotate = 360] [fill={rgb, 255:red, 0; green, 0; blue, 0 }  ][line width=0.08]  [draw opacity=0] (11.61,-5.58) -- (0,0) -- (11.61,5.58) -- cycle    ;

%Straight Lines [id:da1767564447957961] 
\draw    (466,108) -- (466,161.86) ;
%Straight Lines [id:da8139219457300351] 
\draw [line width=1.5]    (466,108) -- (426,108) ;
\draw [shift={(422,108)}, rotate = 360] [fill={rgb, 255:red, 0; green, 0; blue, 0 }  ][line width=0.08]  [draw opacity=0] (11.61,-5.58) -- (0,0) -- (11.61,5.58) -- cycle    ;
%Straight Lines [id:da8740848464197022] 
\draw [line width=1.5]    (466,161.86) -- (426,161.86) ;
\draw [shift={(422,161.86)}, rotate = 360] [fill={rgb, 255:red, 0; green, 0; blue, 0 }  ][line width=0.08]  [draw opacity=0] (11.61,-5.58) -- (0,0) -- (11.61,5.58) -- cycle    ;

% Text Node
\draw (12,28.4) node [anchor=north west][inner sep=0.75pt]  [font=\Large]  {$P_{1}$};
% Text Node
\draw (12,87.73) node [anchor=north west][inner sep=0.75pt]  [font=\Large]  {$P_{2}$};
% Text Node
\draw (12,147.06) node [anchor=north west][inner sep=0.75pt]  [font=\Large]  {$P_{3}$};
% Text Node
\draw (12,206.4) node [anchor=north west][inner sep=0.75pt]  [font=\Large]  {$P_{4}$};
% Text Node
\draw (475,49.4) node [anchor=north west][inner sep=0.75pt]  [font=\Large]  {$TF( P_{1} ,P_{2}) =1$};
% Text Node
\draw (475,124.9) node [anchor=north west][inner sep=0.75pt]  [font=\Large]  {$TF( P_{2} ,P_{3}) =1$};
% Text Node
\draw (475,200.4) node [anchor=north west][inner sep=0.75pt]  [font=\Large]  {$TF( P_{3} ,P_{4}) =0$};

\end{tikzpicture}
}
\caption{Demonstration of how the transformer classifier is used during inference, by comparing consecutive paragraphs to detect section boundaries.}
\label{fig:inference}
\end{figure}
For inference, we use the classifiers of the previous step as topic change detectors for text segmentation. We read each paragraph of the document sequentially and classify the adjacent paragraphs for topical mutuality. More concretely, given a document $d \in D$ divided into consecutive paragraphs $ P=(p_{1},....p_{k})$, section breaks are marked as where the paragraph's topic changes. 
Considering a transformer $TF$ as our classifier and two consecutive paragraphs as our input, the classifier outputs the probability of the two paragraphs belonging to the same topic, independent of their surrounding context, e.g., $TF(p_{1}, p_{2}) = P( Topic(p_{1}) =Topic(p_{2}))$. 
Therefore, given sequences of paragraphs $p_{1},....p_{k}$, and the corresponding predicted labels $y = (y_{1},... , y_{k-1})$, a segmentation
of the document is given by $k - 1$ predictions of $TF$, where $y_{i}=0$ denotes the end of a segment by $p_{i}$. It is worth noting that regardless of the chunk type used during the training of the classifiers (section or paragraph inputs) 
the segmentation module operates on paragraphs only. Figure~\ref{fig:inference} shows the inference on a sample document with four paragraphs and two sections, where the paragraph colors show the topics. The $TF$ classifier is applied on a paragraph pair and can ideally recognize the topic change from $P3$ to $P4$ and mark the beginning of the new section. 

\subsection{Legal Applications}
To put the presented segmentation into a legal context, we focus on three main application areas:
\begin{inparaenum}[$(i)$]
\item As mentioned, a section-based semantic segmentation can be used as a pre-processing step for a passage retrieval context. This, however, would require additional data with relevance annotations for both sentence- and paragraph-level relevance to compare the specific benefits of our approach, which we leave to future work in this area.
\item However, semantically coherent sections can also be used as a basis for similarity search. This is especially helpful when looking for, e.g., related sections in existing contracts \cite{DBLP:conf/jurix/WestermannSWAB20}. Here, we focus on Terms-of-Service documents that are widely available, and contain sections that follow a general pattern of similar topics.
\item Lastly, the section separation can be used for generating outlines of documents, which has previously been shown to work well on other domains such as Wikipedia \cite{DBLP:conf/sigir/ZhangGFLC19}. During our document crawl, we also encountered several documents not including any sectional headings, which makes it especially hard to understand the legal contexts for laymen users.
\end{inparaenum}

%\noindent In the next section, we go over the properties of our Terms-of-Service dataset used in the experiments in Section~\ref{sec:eval}, where we go over three different training strategies for the IID Same Topic Prediction. First, we define the chunks as sections and train our classifiers to detect the sections with the same topics. Second, we use the paragraphs that have the same topic as positive examples and randomly sample for negative examples among the paragraphs with different topics. In this approach, the positive samples could be from paragraphs in different documents. Finally, we restrict our positive and negative sampling strategy to only inter-document instances, to detect inter-document dependencies. We investigate the implication and effect of the various training strategies on the same topic prediction task and text segmentation of a held-out set. 

\section{Terms-of-Service (ToS) Dataset}
\label{sec:tos}
%!TeX root=main_paper.tex

Due to data governance policies in many countries, it is generally mandated that commercial websites contain the necessary legal information for site users. Specifically, these must be easily reachable via the landing page, which makes it comparatively easy to be crawled. For each Terms-of-Service document, we automatically extract the content divided into paragraphs and respective hierarchical section headings. Further, ToS documents allow us to experiment with a large-scale dataset that comes with a shared set of topics, while still maintaining a heterogeneous set of topics due to the different types of websites. In the following, we will discuss the detailed mining process, and implicate limitations of this approach.

\subsection{Crawling}

As seeds to our crawler, we use the Alexa 1M URL dataset.\footnote{Available at: \url{http://s3.amazonaws.com/alexa-static/top-1m.csv.zip}} For each URL in the dataset, we try to access the website both with and without the \texttt{www} prefix. First, the landing page is downloaded and parsed using the \textit{Beautiful Soup} Python package. We then search for hyperlinks with texts \textit{Terms of Service}, \textit{Terms of Use}, \textit{Terms and Conditions}, and \textit{Conditions of Use}, and follow them to get to the respective terms-of-service pages. Levenshtein distance with a threshold of 0.75 is used to allow for spelling mistakes and different wording. %(e.g., \textit{Terms \& Conditions} instead of \textit{Terms and Conditions}). 
The raw Hypertext Markup Language (HTML) content of the Terms-of-Service page is downloaded and stored for further processing. In case of an error, e.g., if the website is temporarily unreachable, we retry the same website 2 additional times before skipping it. The unprocessed dataset contains HTML code for roughly $74{,}000$ websites. Note that due to limitations of the current crawler implementation, websites that rely on JavaScript to display content are not supported. % The idea of using \texttt{Selenium} was discarded due to the significant overhead in crawling time.

\subsection{Section Extraction}

Despite the fact that HTML is a structured format, it is a non-trivial task to extract text and hierarchies. The main reasons are that Web pages often contain a lot of boilerplate (e.g., navigational elements, advertisements, etc.), generally have heterogeneous appearences and implementations, and that they simply do not always conform to the HTML standard.
Here, only a rough outline of the pipeline is given. For further reference, please refer to the implementation in our repository.
%In the following we describe the pipeline used for section extraction. We do not go into all details, but only explain the rough structure of the algorithm. For the details, please refer to the reference implementation. The algorithm consists of the following steps:

\subsubsection*{Boilerplate Removal}
For boilerplate removal, we use the \texttt{boilerpipe} package by Kohlsch\"utter et al.~\cite{DBLP:conf/wsdm/KohlschutterFN10}, which is based on shallow text features for classifying the text elements on a Web page. The result is an HTML page with all navigational elements, advertisements, and template code removed. Importantly, relevant hierarchical information is retained past this step. % Most importantly though, the structure we are interested in is preserved in this step.

\subsubsection*{HTML Cleanup}
To deal with websites that do not conform to HTML standards, we perform several cleanup steps. This includes, for example, fixing mistakes such as text appearing without a corresponding paragraph (\texttt{<p>} tag), or incorrectly nested tags (e.g., section headings within a \texttt{<p>} tag). We fix such mistakes by adding missing tags and adjusting nested tags similar to how a web-browser would interpret the code.

\subsubsection*{Language Detection}
Since the Alexa dataset also contains many non-English websites, we reject extracted terms-of-service, where the majority of text most likely has a language different from English. We use the \texttt{langid} Python package for detecting the language of each individual paragraph (\texttt{<p>} tag).

\subsubsection*{Extracting Hierarchy}
To obtain the hierarchy, we split the document into smaller chunks. Splits are done in the following order: first we split on each section heading (\texttt{<h1>}-\texttt{<h6>} tags), then on bold text (\texttt{<b>} tag) starting with an enumeration pattern, then on enumerations (\texttt{<li>} tags), then on underline text (\texttt{<u>} tag) starting with an enumeration pattern, and lastly on regular text (\texttt{<p>} tag) starting with an enumeration pattern. To prevent spurious splits, each criterion is only used if there are at least 5 occurrences within the document. Each time the document is split, we save the corresponding headings, which then form the hierarchy. As enumeration patterns we recognize Latin numbers, roman numerals, and letters, optionally prefixed with \textit{Part}, \textit{Section}, or \textit{Article}.
The majority of documents contain at most two levels of section hierarchy.

\subsection{Data Set Statistics}

In addition to the full dataset, we provide a cleaned subset for which we manually grouped sections into distinct topics based on similar spelling or meaning. We manually merged $554$ section titles, which corresponds to all titles with at least $250$ distinct occurrences in the corpus. After merging, $82$ topics were obtained, and only sections that have at least one of these aliases as a heading were kept.
The dataset contains different levels of section hierarchy. For our work, we group document content into top-level sections only, any further hierarchies are discarded, but are present in the raw data and available for future work. After removing documents without any valid sections, belonging to the predefined $82$ topics, we are left with approximately $43{,}000$ documents for the same section prediction task, and around $40{,}000$ documents for our paragraph-level setup.
 We randomly split the data $80/10/10$ into train, validation, and test set. The average number of sections per document is $6.56$, and each document consists of $22.32$ paragraphs on average, which results in a mean of $2.92$ paragraphs per section.
\Cref{stats} shows the top 10 section topic labels. The average number of paragraphs per section varies between different topics.

\begin{table}[t]
\caption{Top 10 section topics by document frequency. Additionally, the number of associated paragraphs is given.}
\label{stats}
\setlength{\tabcolsep}{6pt} % Default value: 6pt
\renewcommand{\arraystretch}{1} % Default value: 1

\begin{tabular}{lrr}
\toprule
Topic Label & \pbox{20cm}{Document \\ Frequency} & \pbox{20cm}{Paragraph \\ Frequency}  \\
\midrule
limitation of liability & $21{,}317$ & $68{,}517$\\ 
indemnification & $16{,}698$ & $25{,}683$ \\
law and jurisdiction & $15{,}113$ & $29{,}790$ \\
links to other websites & $13{,}752$ & $24{,}727$ \\
termination & $12{,}855$ & $33{,}978$ \\
warranty & $9{,}926$ & $41{,}403$ \\
privacy & $8{,}958$ & $25{,}022$ \\
disclaimer & $8{,}575$ & $29{,}265$ \\
general terms & $7{,}936$ & $54{,}693$ \\
\bottomrule
\end{tabular}%

\end{table}

\section{Evaluation }
\label{sec:eval}
%!TeX root=main_paper.tex
\begin{figure}
\centering 
\resizebox{0.5\textwidth}{0.43\textwidth}{      
\input{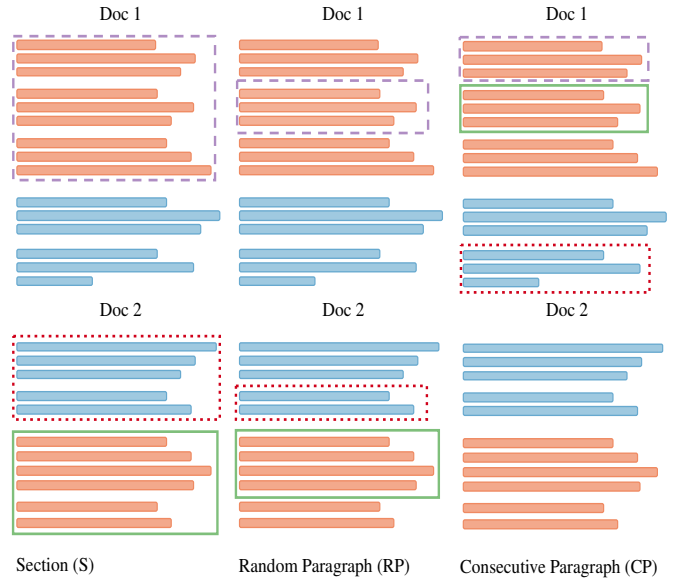}
}
\caption{Visualization of three distinct setups for same topic classification, where the example chunk of text is depicted in dashed purple, a positive sample in line of green, and a negative sample in dotted red. Each vertical line is a sentence and the grouping of them shows a paragraph, different sections in the document are shown with different colors, denoting the same topic across all paragraphs of the same section. From left to right: Same Section prediction, Random Paragraph, and Consecutive Paragraph sampling.}
\label{fig:sampling}
\end{figure}

\begin{table*}[t]
\caption{Prediction accuracy for the independent topic prediction tasks, Same Topic Prediction (STP), Random Paragraph (RP), Consecutive Paragraph (CP) with different sampling strategies. Standard deviation is reported over 5 runs and the best model on each respective set is depicted in bold.}
\label{SSPtab}
\setlength{\tabcolsep}{10pt} % Default value: 6pt
\renewcommand{\arraystretch}{1} % Default value: 1
\resizebox{\textwidth}{!}{%
\begin{tabular}{llcccccccc}
\toprule
 && GLV$avg$ & tf-idf & BoW && Ro-CLS && ST-Ro & ST-Ro-N \\
 \cmidrule(lr){3-5}
 \cmidrule(lr){7-8}
 \cmidrule(lr){9-10}
\multirow{2}{*}{STP} & Dev & 89.70 \rpm .07 &82.10 \rpm .05 & 50.94 \rpm .33 && 96.42 \rpm .52 && 96.38 \rpm .03 & \textbf{96.39 \rpm .03}\\ 

   & Test & 90.01 \rpm .06 & 82.54 \rpm .07 &51.05 \rpm .51 && 96.58 \rpm .52 && 96.45 \rpm .06  & \textbf{96.46 \rpm .02}\\
 
\midrule

\multirow{2}{*}{RP}   & Dev & 76.63 \rpm .04 & 70.94 \rpm .07 & 50.34 \rpm .04&& 57.63 \rpm 10.4 && \textbf{87.50 \rpm .13}  & 87.39 \rpm .08 \\
   & Test & 76.16 \rpm .06 & 70.41 \rpm .09 & 50.31 \rpm .37 && 57.48 \rpm 10.2 && \textbf{87.19 \rpm .64} & 86.88 \rpm .11  \\

 \midrule
\multirow{2}{*}{CP}  & Dev & 77.64 \rpm 6.6 & 74.94 \rpm .11 & 56.34 \rpm .83 && 89.63 \rpm .12 && \textbf {91.17 \rpm .05} & 91.12 \rpm .04  \\
 & Test & 78.63 \rpm 6.8 & 76.17 \rpm .07 & 56.58 \rpm 1.1 && 90.34 \rpm .08 && 91.17 \rpm .04   & \textbf{91.69 \rpm .02}\\
\bottomrule
\end{tabular}%
}
\end{table*}

We demonstrate the capabilities of transformer-based architectures for topical change detection using dataset consisting of online Terms-of-Service (ToS) documents, which was discussed above. Results are compared for the introduced IID STP task (see \Cref{sec:STP}) as well as a downstream comparison of text segmentation results to a range of baselines and existing methods. Results show a great improvement in the performance for all transformer-based models.

\subsection{Evaluation of Models}
\sloppy
We compare our methods against a range of baselines, including averaging over Global Vectors (\textit{GLVavg})~\cite{DBLP:conf/emnlp/PenningtonSM14}, tf-idf vectors (\textit{tf-idf}), and Bag of Words (\textit{BoW})~\cite{Zellig1954}. 
For transformer language models, we evaluate the standard \texttt{[CLS]} sequence classification with \texttt{roberta-base} (\textit{Ro-CLS}). For Sentence-Transformers \cite{DBLP:conf/emnlp/ReimersG19} we use the Siamese transformer setup with a variant of \texttt{roberta-base} (\textit{ST-Ro}) and an additional  model that has been pre-trained on NLI sentence similarity tasks (\textit{ST-Ro-NLI}) to investigate performance of further pretraining.

\noindent Transformer models are trained using the popular HuggingFace transformer library \cite{Wolf2019HuggingFacesTS} for the \texttt{[CLS]} models, as well as the sentence-transformers package from \cite{DBLP:conf/emnlp/ReimersG19} for Siamese variants. We use two Nvidia Titan RTX GPUs for training, and each model variant has been trained with five different random seeds. Details for the training parameters can be found in our public repository. Due to the length limitation of $512$ tokens, we employ an iterative truncation strategy for two-sentence inputs. Due to the coupling of two transformers for the  Sentence-Transformers the input size doubles, accepting a total input of $1024$ tokens.

\subsection{Prediction Tasks}
As previously introduced, we train models with an independent classification setup, which is generally much faster than more complicated hierarchical sequential models. Specifically, we highlight the differences in the setup for the same section prediction task, compared to the two paragraph-based methods. We point out that results depicted in \Cref{SSPtab}, for the prediction accuracy are not directly comparable between sampling methods, as they generate different development and test sets based on the employed sampling strategy. We show in the subsequent section, however, that downstream performance for the text segmentation is in line with results on topic prediction. Figure~\ref{fig:sampling} visualizes the different sampling strategies. In the following, we describe each strategy in detail and highlight their difference. Across all strategies, we added three positive and three negative samples for each individual section/paragraph.

\subsubsection{Section (S) Topic Prediction.}
In this setup, we use sections as input chunks to the transformer classifiers.
The section task showcases how different levels of granularity can affect outcomes in the prediction results. Specifically, the extremely long input sequences test the limits of what transformers can predict from partial observations since the majority of inputs will be heavily truncated. 
To ensure an equal distribution of samples from within the same and different sections, we match each section with three samples from the same topic, and three from different topics. The positive and negative sections can be sampled from a different document. The important point is that the positive samples should come from the \emph{same topic} and negative samples from different ones. The first column of Figure~\ref{fig:sampling} visualizes the section sampling, where the first section of $Doc_1$ is paired with the second section of $Doc_2$ to form a positive sample and the first section of $Doc_2$ to form a negative sample, respectively. The same strategy is employed for the generation of the development and test set.

\noindent Despite the constraints with respect to the input length, we find that all transformers perform on a near-perfect level, compare \Cref{SSPtab}. Comparing these results to already very well-performing baselines, we suspect that certain keywords give away similar sections, but highlight the fact that the explicit representation of different topics is not given during training in the binary classification task, which makes this a suitable method for dealing with imbalanced topics.

\subsubsection{Random Paragraph (RP) Topic Prediction.}
In contrast to the section-level task, we revert to a more fine-grained distinction of paragraphs in a text. In the Random Paragraph setting, we still generate samples similarly, meaning we include three paragraphs from a random document with the same topic and three negative samples from random paragraphs with different topics. The main difference between the Section prediction and Random Paragraph is in the level of granularity and not how the samples are chosen. The second column of Figure~\ref{fig:sampling} highlights this difference, where the samples are paragraphs inside the sections rather than the entire section. Paragraph-based sampling is closer to our inference setup, where each input document is considered one paragraph at a time. However, results show a sharp drop in the performance, which can come from a much narrower context of the paragraphs, as well as a differing selection of test samples compared to the section task. Solely the BoW model seems to be largely unaffected, which is simply due to its low performance in either setting.

\subsubsection{Consecutive Paragraph (CP) Topic Prediction.}
To boost performance and account for the coherent structures in the text, we employ a sampling strategy inspired by Ein Dor et al.~\cite{ein-dor-etal-2018-learning}. For their triplet loss, samples are generated inside the same document only, which can be translated into sampling from intra-document paragraphs. Note that this strategy also no longer requires any merging and annotation of topics across documents, as all relevant information is now contained within a single document. This fact opens up much larger generation of training data, which we omit in our current work for the sake of comparability with the RP model.
To generate samples, we look at all paragraphs of a section and pair them as positive samples. Negative samples are picked from paragraphs of different sections in the same document. The third column of Figure~\ref{fig:sampling} depicts the consecutive paragraph setup, where the samples are limited to paragraphs of $Doc1$. Note that despite their similar setup, results of RP and CP runs in \Cref{SSPtab} are not evaluated on the same test set and thus are not comparable, since the test sets are each generated with the respective sampling strategies (RP or CP) as well. However, we are able to compare their downstream performance on the subsequently introduced text segmentation task (see \Cref{sec:textseg} and \Cref{OGtab}). \\

\noindent The result of different sampling strategies along with the performance of the baselines is shown in Table \ref{SSPtab}, where the transformer-based models all outperform the baselines by a significant margin. Among the baselines BoW has the worst performance overall, with the accuracy close to random, showing that distinct word occurrences are not a sufficient indicator. Average GloVe has the best performance of all baselines, but is still behind the transformers by a large margin. Despite the NLI-pretrained SRoBERTa model (ST-Ro-N) achieving better scores than the base model (ST-Ro) for most setups, the difference is insignificant, indicating that the pre-training on sentence similarity tasks does not directly influence our topic prediction setup.
\begin{table}[t]
\caption{Boundary error rate $P_k$ for compared models (lower is better), based on sampling strategies Random Paragraph (RP), Consecutive Paragraph (CP) and their Ensemble variates, RP$_{Ens}$ and CP$_{Ens}$, respectively. Ensemble (''\textit{Ens}'') predictions are obtained by majority voting over model runs.}
\label{OGtab}
\setlength{\tabcolsep}{6pt} % Default value: 6pt
\renewcommand{\arraystretch}{1} % Default value: 1
\resizebox{0.5\textwidth}{!}{%
\begin{tabular}{lccccc}
\toprule
 & RP  & CP && RP$_{Ens}$ &  CP$_{Ens}$  \\
\midrule
\midrule
 GLV$avg$ & 29.97 \rpm .09  & 26.23 \rpm 6.2 && 29.55 & 23.06 \\ 
 tf-idf & 39.87 \rpm .24 & 29.70 \rpm .28 && 39.36 & 28.60 \\
  BoW & 45.76 \rpm .67 & 43.46 \rpm 1.5 && 46.20 & 41.80 \\
 Random Oracle & 35.08 \rpm .15 & - && 31.88 & - \\
 \midrule
 
GraphSeg & - & 32.48 \rpm .46 && - & 32.28 \\
WikiSeg  & - & 48.29 \rpm .30 && - & 48.29 \\
\midrule
 %\midrule(lr){1-1}
 %\cmidrule(lr){2-3}
 %\cmidrule(lr){5-6}

 Ro-CLS & 37.26 \rpm 4.8 & 15.15 \rpm .00 && 41.15 & 15.15 \\
 \midrule
 %\cmidrule(lr){1-1}
 %\cmidrule(lr){2-3}
% \cmidrule(lr){5-6}
 ST-Ro & 15.72 \rpm .11 & 14.06 \rpm .14&& \textbf{14.62} & 13.14 \\
ST-Ro-N & 15.97 \rpm .14 & 13.97 \rpm .19 && 14.81 & \textbf{12.95} \\

\midrule
Ens consec & - & - && - & 12.50 \\
\bottomrule
\end{tabular}%
}
\end{table}

\begin{figure}[t]
	\centering
	\includegraphics[width=0.47\textwidth]{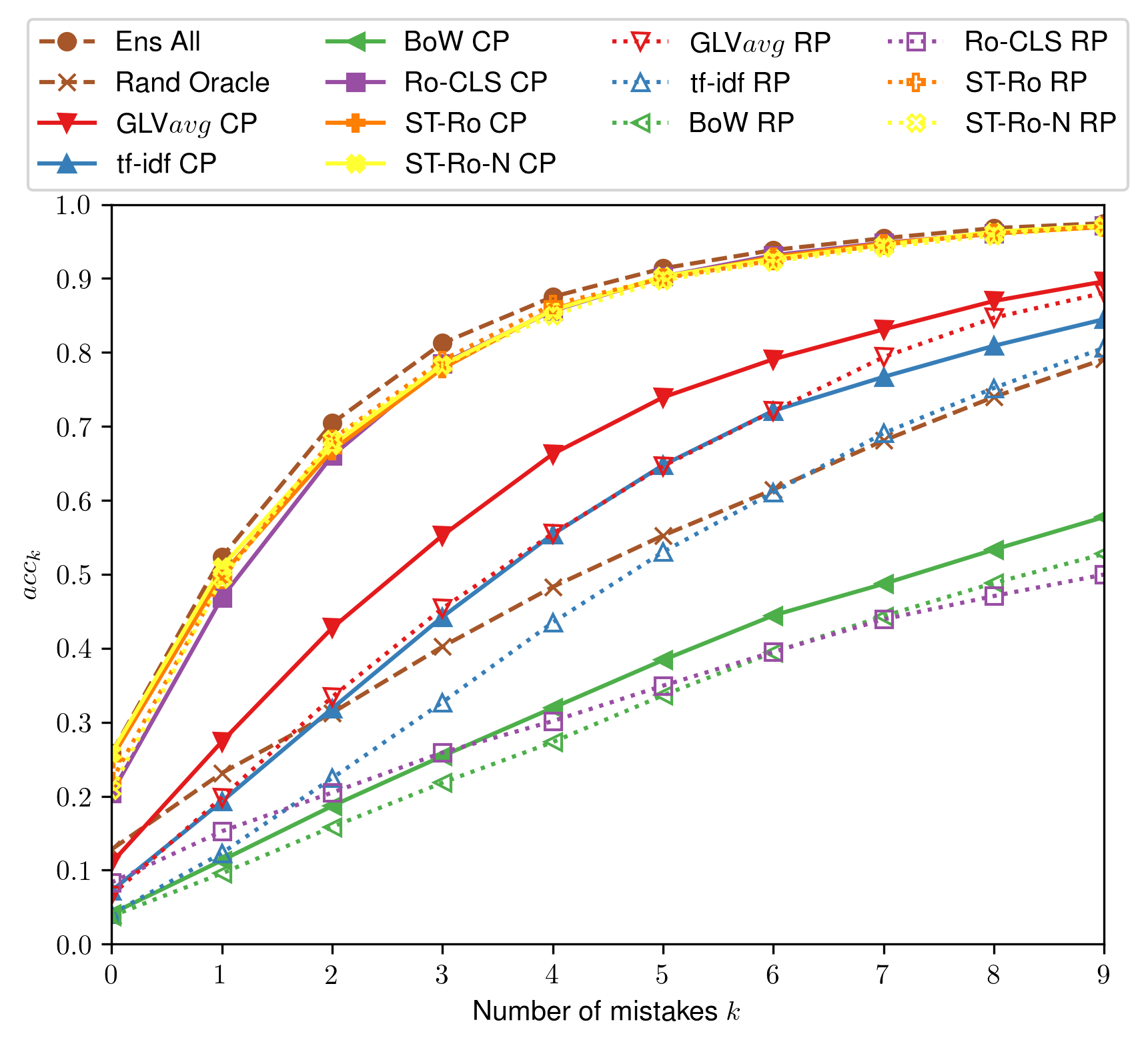}
	\caption{Mistake rate of per-model ensembles, where the suffix CP indicates the consecutive paragraph sampling and RP the random paragraph sampling for each model. The baseline is Rand Oracle (Random Oracle), GLV$avg$ (average GloVe vectors), tf-idf, and BoW (Bag-of-Words). Ro-CLS is the fine-tuned CLS token for Roberta and Sentence transformer models are ST-Ro and ST-Ro-N, where the latter is pre-trained on NLI task. The best performing model is the Ens All (Ensemble of all models).  }
	\label{OGfig}
\end{figure}
\subsection{Text Segmentation}\label{sec:textseg}

By generating a text segmentation over the paragraphs of a full document, the independent prediction results from the previous section can now be compared across several approaches. Specifically, we compare the paragraph-based training methods CP and RP.
\noindent As an evaluation metric, we follow related literature and adopt the $P_k$ metric introduced by Beeferman et al.~\cite{DBLP:journals/ml/BeefermanBL99}, which is the error rate of two segments at $k$ sentences apart being classified incorrectly. We use the default window size of half the document length for our evaluation, again following related work.
Furthermore, we count the number of explicit misclassifications, and use the accuracy $acc_k$ of ``up to $k$ mistakes per document'' as an evaluation metric.
Due to the coarser nature of paragraphs and the lower number of predictions per document compared to the sentence-level segmentation, this is a more illustrative metric. This also relates to the ``exact match'' metric $\mathbf{E}\mathbf{M}_{outline}$ employed by Zhang et al.~\cite{DBLP:conf/sigir/ZhangGFLC19}, where $acc_0 = \mathbf{E}\mathbf{M}_{outline}$. \\
%For the comparison to other baselines, especially,  considering the sequential nature of the text segmentation task, we were unfortunately not able to obtain functioning source code or change the respective implementations to work with the relaxed paragraph/section setup.
\noindent Here, we also include the performance of related works where public and up-to-date code repositories are available. Specifically, we compare to the unsupervised segmentation algorithm GraphSeg~\cite{DBLP:conf/starsem/GlavasNP16}, and the supervised model by Koshorek et al.~\cite{DBLP:conf/naacl/KoshorekCMRB18}, which we dub ``WikiSeg''. Both approaches are trained on a sentence-level approach, though, and predictions have to be translated back to a paragraph level for comparison of results. We train each model with the suggested parameters in their publicly available repositories.
For an additional pseudo-sequential baseline, we use an informed random oracle that has a-priori information on the number of topics in the document, and samples from a distribution with adjusted probability $P(\text{``next section''}) = \#sections / \#paragraphs$. 
Note that no additional parameters are learned for any model, and predictions are binarized with a simple $0.5$ threshold over the same topic predictions. 
We provide ensembling results for the majority voting decisions by the five seed runs of each model variant (\textit{Ens}), which provides further improvements. Best results are obtained by ensembling all consecutive transformer-based methods (\textit{Ens consec}).\\
\Cref{OGtab} shows the results of the evaluation, where one can see that results in the sequential segmentation are directly linked to the performance on the independent classification task seen in \Cref{SSPtab}.  To verify our initial assumption of cross-document comparability of content from similar sections, we make the following observations: 
\begin{inparaenum}[$(i)$]
\item Evaluation performance for the STP setup is consistent for both training strategies (RP and CP) when using Sentence-Transformer models (see \Cref{SSPtab}). 
\item Similarly, both CP and RP-trained Sentence-Transformer segmentations achieve results within 2 percentage points of the respective $P_k$ scores.
\item In general, CP training setup yields slightly better $P_k$ scores, likely because the intra-document dependencies are captured better with this sampling strategy, which is a more appropriate sampling for the segmentation task.
\item We find convergence problems for RP training with the \texttt{[CLS]} models, as well as the tf-idf model.
\end{inparaenum}
Due to the size of our general training corups, we therefore conclude that it is realistic to expect topical similarity within a section, \emph{even across documents}. However, due to the seemingly inconsistent convergence of RP models, we caution against blindly using this strategy, especially when dealing with more heterogeneous corpora.
The oracle baseline performs unexpectedly better than both tf-idf and BoW, indicating that additional information about the sections of a document can greatly boost task performance, which might be relevant for future work.
Additional pre-training of ST models (ST-Ro-N) does not show any significant improvement over the standard ST-Ro models.\\

\noindent To our surprise, sentence-based implementations (GraphSeg and WikiSeg) show significantly lower performance, and fall even behind the simpler baselines.
For GraphSeg, an unsupervised segmentation approach, the lack of explicit training on the different granularity seems to significantly prohibit correct predictions on longer segments.
WikiSeg heavily preprocesses the data and discards many samples, thus significantly shrinking the training set.
Since performance on the reduced training set is still decent, this indicates that training a network from scratch is not suitable with the smaller training set of a reduced corpus and tends to overfit. We expect a significant increase in performance if the training would instead be performed without such strict preprocessing criteria, or continuing fine-tuning on pre-trained weights from a paragraph-level WikiSeg model.
For either baseline model, it is also important to note that these models predict on the entirety of the sequence, which theoretically allows information sharing between different sections in the current sample. However, they show no improvement over our binary prediction setup which does not share this information. It would be of interest to compare results to sequential transformer-based architectures, such as they are used by Glavas et al.~\cite{DBLP:journals/corr/abs-2001-00891}. However, their model again requires training from scratch, which has proven to be inconsistent in our experiments with WikiSeg.\\

\noindent Lastly, the plots for $acc_k$ for various models in \Cref{OGfig} indicate a correlation between the $acc_k$ and $P_k$ measures, which does not apply to sentence-level segmentations. Overall, the best-performing ensembles classify around $25$\% of documents without any mistake ($acc_0$), and around $70$\% with less than three mistakes ($acc_2$) over the entire document. We therefore suggest $acc_k$ as an interpretable addition to the classic evaluation of segmentation approaches when dealing with paragraph-level segmentations.

\section{Conclusion and Future Work}
\label{sec:conclusion}
%!TeX root=main_paper.tex

Despite a multitude of previous works, structural text segmentation methods have always focused on very finely segmented text chunks in the form of sentences. In this work, we have shown that a relaxation of this problem to coarser text structures reduces the complexity of the problem, while still allowing for semantic segmentation.
Further, we reformulate the oftentimes expensive-to-train sequential setup of text segmentation as a supervised Same Topic Prediction task, which reduces training time, while allowing for a near-trivial generation of samples from automatically crawled text documents.
To show the applicability of our method, we present a new domain-specific and large corpus of online Terms-of-Service documents, and train transformer-based models that vastly outperform a number of text segmentation baselines.

\noindent We are currently investigating the setup for deeper hierarchical sections, which our dataset already contains annotations for, to see whether such notions can also be picked up by an independent classifier and benefit a legal retrieval system. Also, the findings from our Consecutive Paragraph model already indicate that training requires no further information than the ground truth segmentation, which can generally be inferred from structured input formats, such as HTML or XML, making this an attractive option for a larger-scale study of cross-domain document collections.
Finally, an interface build on top of our framework, enabling the users to judge the usefulness of segmentation for legal use cases, such as a collection of documents from mergers and acquisitions, could be used to determine the efficacy of our improved segmentation.

\section*{Acknowledgements}
We thank the anonymous reviewers for their insightful comments.

 \bibliographystyle{ACM-Reference-Format}
  \bibliography{icail_paper_authors_version}
\end{document}